\pdfoutput=1
\documentclass[11pt]{article}

\usepackage{EMNLP2023}
\usepackage{tabularx}
\usepackage{booktabs, multirow} % for borders and merged ranges
\usepackage{soul}% for underlines
\usepackage{float}
\restylefloat{table}

\usepackage{times}
\usepackage{latexsym}
\usepackage{tikz}
\usepackage{amsfonts}
\usepackage{xspace}
\usetikzlibrary{topaths}
\tikzstyle{every picture}+=[remember picture,inner xsep=0,inner ysep=0.05ex]

\usepackage[T1]{fontenc}

\usepackage[T1]{fontenc}    % use 8-bit T1 fonts
\newcommand\funfont[1]{{\usefont{T1}{QTWestEndRegular}{m}{n}#1}}
\newcommand{\dataname}{\funfont{CompPrompts}\xspace}
\newcommand{\evalname}{\funfont{ControlledImCaps}\xspace}
\usepackage[utf8]{inputenc}

\usepackage{microtype}

\usepackage{inconsolata}

\title{
Text encoders bottleneck compositionality\\ in contrastive vision-language models
}

\author{
Amita Kamath$^1$ \hspace{2em} Jack Hessel$^2$ \hspace{2em} Kai-Wei Chang$^1$\\
$^1$ University of California, Los Angeles \\
$^2$ Allen Institute for AI \\
\texttt{\{kamatha, kwchang\}@cs.ucla.edu}, \texttt{jackh@allenai.org}
}

\begin{document}
\maketitle
\begin{abstract}

Performant
vision-language (VL) models like CLIP % excel at detection and retrieval %have proven successful
% despite the fact that they
represent captions using a single vector.
%introduce representational bottlenecks for efficiency, e.g.,
% to support large contrastive batch sizes,
%and multimodal interactions are captured via a simple dot product.
%How much information about language is lost under this transformation? 
How much information about language is lost in this bottleneck?
 % This raises the question, well-studied in NLP, of how much information you can pack into one embedding. --- maybe for intro we can include the quote in a footnote :-)
%\amita{Is there some short form for CLIP-like VL models?}
%% JMH ---I moved this to a potential intro :-)
% We posit that if you can't reliably decode a specific property (e.g. spatial relations) from such a VL model's text embeddings, the VL model's text encoder does not encode it, and therefore the VL model \textit{cannot}.
% Importantly, our method, in contrast to prior work, is text-only, which allows fine-grained testing and is not restricted by images that currently exist.
We first curate \dataname, a set of increasingly compositional image captions that VL models \emph{should} be able to capture (e.g., single object, to object+property, to multiple interacting objects).
Then, we train \emph{text-only recovery probes} that aim to reconstruct captions from single-vector text representations produced by several VL models.
% of them generated by different VL models.
%% lets cut this --- we dont need to just argue anything anymore, we have numbers!
% We argue that if a textual property cannot be decoded from the vector representation with a highly expressive decoder, then it cannot be readily modeled in the multimodal setting. 
This approach does not require images, allowing us to test on a broader range of scenes compared to prior work.
We find that: 1) CLIP's text encoder falls short on more compositional inputs, including object relationships, attribute-object association, counting, and negations;
%2) --- flaws that are known to plague VL models, but that until now were not known to be \amita{sufficiently?} due to the text encoder.2)  --- this can probably go in related work; stanford paper was fairly recent, so folks might not know this...
2) some text encoders work significantly better than others; and
3) % even the best encoders still have headroom. 
text-only recovery performance predicts multimodal matching performance on \evalname:
% Finally, we curate
a new evaluation benchmark we collect and release consisting of fine-grained compositional images and captions. % image-and-text with prompt types from \dataname, and verify that on prompt types where the text-only encoders fail, so too do the overall models.
Specifically, our results suggest text-only recoverability is a necessary (but not sufficient) condition for modeling compositional factors in contrastive VL models. % : if the proposed recovery probe cannot
We release our datasets and code.
% \amita{There's tension between "we don't need images" and "we curate image-text pairs"...}

%as single vectors, and then, to quantify information loss, attempt to decode  

%Decode probe mmeasures what 
%We examine top-performing CLIP-like VL models and show that the text encoder can serve as an \amita{informational? whatever's in the title} bottleneck, necessarily resulting in poor or un-grounded predictions. 
%The VL text encoders perform especially poorly on spatial relations, counting, and attribute-object association --- flaws that are known to plague VL models, but that until now were not known to be \amita{sufficiently?} due to the text encoder.
% Something about how improving the text encoder isn't sufficient but is necessary to get better predictions.
% We show that training with modified losses improves the ability of VL text encoders to encode these properties \amita{CLIP-BoW}, but that even within the one-text-embedding framework, there is significant room for improvement to match NLP-trained sentence embeddings.
\end{abstract}

\section{Introduction}
\emph{``A penguin on Mars wearing a spacesuit and walking a robot dog next to Santa Claus.''} \citet{marktweet}'s text-to-image query is the type that modern multimodal models \emph{should} be able to support. It is spatially precise (the dog is \emph{next to} Santa, not in front), compositional (\tikz[baseline=(node1.base)]\node (node1)  {robot}; \tikz[baseline=(node2.base)]\node (node2) {dog};, but not \tikz[baseline=(node3.base)]\node (node3)  {robot}; \tikz[baseline=(node4.base)]\node (node4) {Santa};), and imaginative (it is unlikely such an image exists already). \begin{tikzpicture}[overlay]
    \draw[-latex] (node1.north) to[bend left=18] (node2.north);
    \draw[-latex] (node3.north) to[bend left=18] (node4.north); \end{tikzpicture}
However, several recent works have shown that a variety of multimodal models (despite achieving strong benchmark performance) are frequently unable to reason about even simple spatial relations or attribute attachments \cite{gokhale2022benchmarking,thrush2022winoground,yuksekgonul2022and}.

Underlying several popular multimodal models like CLIP \cite{radford2021learning}, 
DALL-E 2 \cite{ramesh2022hierarchical} 
% \jack{can we triple check that the paper version of DALLE-2 is pooled repr?}
% , Imagen \cite{saharia2022photorealistic}, 
and ALIGN \cite{Jia2021ScalingUV} is a \emph{pooled text encoder,} i.e., a text representation model that outputs a single vector for a given input caption.\footnote{Pooled text encoders (c.f., bidirectional multimodal encoders) are used for a variety of practical reasons: e.g., for guided diffusion \cite{dhariwal2021diffusion}, for fast k-NN queries over billions of images \cite{LAION}, for contrastive objectives dependent on large batch sizes like \newcite{radford2021learning}'s 32K example ``mini''-batch, etc.}
In this work, we use this representational bottleneck as an interface to ask: how precise are textual representations of visually-descriptive language in these modern multimodal models?

\begin{figure}
    \centering
    \includegraphics[width=0.44\textwidth]{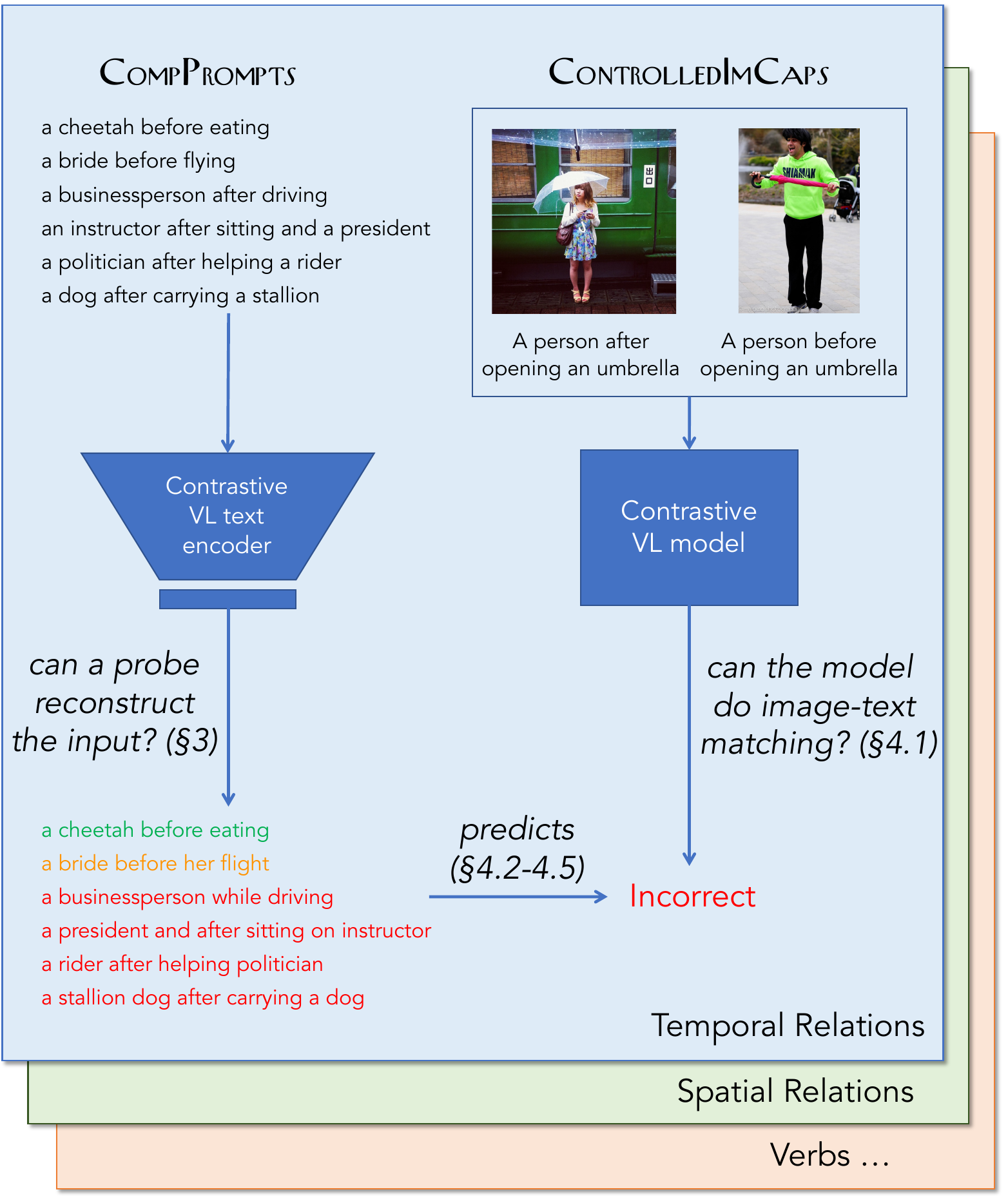}
    \caption{
    We present \dataname, a dataset of 18,100 text prompts, and \evalname, a dataset of 600 image pairs+captions that differ by only one word.
    The two datasets are grouped by the same set of caption properties, e.g., temporal/spatial relations.
    Experiments on \dataname quantify the information loss of a text encoder; experiments on \evalname illustrate that information loss correlates with multimodal errors.
    % We also show in a separate evaluation that this behavior continues downstream.
    }
    \label{fig:teaser1}
\end{figure}

\begin{figure}
    \centering
    \includegraphics[width=0.4\textwidth]{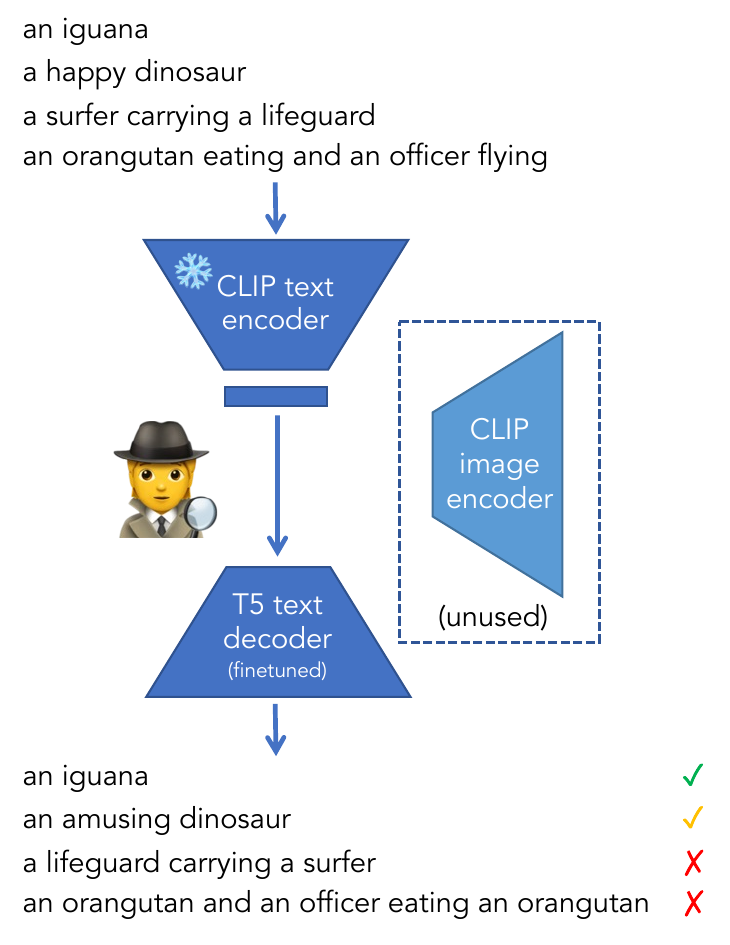}
    \caption{We probe the representations of single-vector text encoders used in popular VL models. Using a corpus of increasingly compositional image captions, \dataname, we attempt to generatively decode the original input sentence. Text encoders of popular models like CLIP fail to effectively encode precise aspects of their captions like attribute attachments and object relationships (real examples shown, as in Figure \ref{fig:teaser1}). 
    % We also show in a separate evaluation that this behavior continues downstream.
    }
    \label{fig:teaser}
\end{figure}

Our strategy is as follows: for a fixed pooled text encoder $\mathcal{T}: x \rightarrow y$, which maps from captions $x$ to vectors $y \in \mathbb{R}^d$, we test how accurately $x$ can be recovered by an expressive generative decoder given $y$, i.e., $\mathcal{P}(x | \mathcal{T}(x))$. In an ideal case, $\mathcal{T}$ should result in \emph{no} information loss, i.e., an exact reconstruction of the original caption should be possible, to account for specific visual factors. 
However, we hypothesize that if a specific visually descriptive property (e.g., a spatial relation) cannot be accurately decoded from $y$ (using a decoder trained with significant supervision), then it is unlikely a multimodal model can effectively use that property of $x$ using $\mathcal{T}$. %, if the model bottlenecks its text representation with $\mathcal{T}$. 
% \amita{However, we show that several specific visually descriptive properties (e.g. spatial relations) can't be accurately decoded from $y$ (using a decoder trained with significant supervision), making it unlikely that a multimodal model can effectively use that property of $x$, if the model bottlenecks its text representation with $\mathcal{T}$.}
Different from existing probes, ours does not require images, % to be associated with captions, 
enabling exploration of a broader range of captions, e.g., creative text-to-image queries for which there may be no associated image (like \emph{``A penguin on Mars...''}).

We execute our probe using an increasingly-compositional hierarchy of image captions we curate, \dataname, which covers cases ranging from a single object with no attributes (e.g. ``a cat'') to multiple objects with attributes and relations (e.g. ``an orange cat to the left of a dog''). We also test counting (e.g. ``three cats'') \cite{segui2015learning,parcalabescu2020seeing} and negations (e.g. ``a cat that is not yawning'').
We compare five text encoders, and find that top contrastive VL models: (1) are broadly ineffective at textually encoding spatial relations, numbers, and negations; (2) frequently cannot match attributes to their corresponding objects; and (3) fail more as inputs grow more compositional. While some text encoders perform significantly better than others, all underperform a proof-of-concept model which demonstrates that our prompts can indeed be compressed into single vectors with little information loss. 

In order to verify that our text-only probe predicts performance in multimodal settings, % claim that information lost by a VL text encoder cannot be accessed by the multimodal model,
we curate an evaluation set of image-caption pairs, \evalname, which operationalizes the compositional factors of \dataname in a multimodal setting. Results on this corpus suggest our text-only probe gives a \textit{necessary} condition: %show that on types of prompts where the
if the text-only recovery probe fails to recover a text-only property on \dataname, then the associated multimodal model also performs poorly for that property on \evalname.  % further verifying that the text encoder bottlenecks model performance. 
However, our results also suggest that text-only recoverability is not a \emph{sufficient} condition: a model can achieve low text-only information loss on a particular prompt type but not fully solve it on \evalname.
% our proof-of-concept encodings, which achieve low text-only information loss, do not fully solve \dataname either.
To facilitate future probing experiments, we release our code alongside the newly collected \dataname and \evalname corpora at \url{https://github.com/amitakamath/vl_text_encoders_are_bottlenecks}.

\section{Evaluation Corpora}
\subsection{\dataname}
We create an evaluation dataset of 18,100 text prompts describing potential visual scenes with varying degrees of specificity and composition.
Our starting point is animate nouns with corresponding verbs and adjectives from the Web10K dataset \cite{kamath2022webly}. We remove some synonyms to prevent ambiguity in the prompt (e.g. ``a rhino to the left of a rhinoceros'').

%\jack{I wonder if there's a fun way of presenting this; ill think about figure 1...}
The prompts are increasingly compositional:
They have 1 or 2 unique nouns, and 0, 1, or 2 attributes, of which there are 4 types: adjective, verb, spatial, and temporal. 
Nouns are randomly matched to generate prompts with two unique nouns --- this results in unusual and imaginative text inputs that cannot be guessed based on priors learned during model pre-training (e.g., ``a crab lifting a rhino'').
The verb and spatial attributes can have either one associated noun (i.e. intransitive, e.g. ``a koala yawning'', ``a policeman on the left'') or two (i.e. transitive, e.g. ``a poet chasing a rabbit'', ``a dinosaur left of a tiger''). 
We also test multiples and negations in the one-attribute setting. 

Prompt examples of each type are given in Tables~\ref{tab: all_em} and \ref{tab: all_negations_multiples}. There are 300-500 examples of each prompt type in the dataset.

\subsection{\evalname}
\label{sec:evalname}
We create a second evaluation dataset to evaluate the overall vision-language model, rather than the text encoder specifically: where \dataname contains text prompts alone, \evalname contains 600 pairs of images, along with a corresponding caption for each image. 

The images are sourced from the COCO validation set \cite{lin2014microsoft}, and the captions are hand-written to study one of six specific fine-grained textual properties: spatial relations with one associated noun, spatial relations with two associated nouns, temporal relations, verbs with one associated noun, verbs with two associated nouns, or adjectives. For spatial relations, we evaluate only ``left'' and ``right'' (unlike \dataname, which evaluates also ``above'', ``under'', ``in front of'', and ``behind''), due to insufficient presence of other spatial relations clearly depicted in the COCO data. 

A key property of \evalname is that \textit{only one word changes} between the two captions associated with a given image pair, such that the relation is changed or inverted: e.g., the caption pair ``a person before opening an umbrella'', ``a person after opening an umbrella'', along with the corresponding images for each (as in Figure \ref{fig:teaser1}) tests the overall model's understanding of temporal relations alone, without conflating any other types of reasoning. 

% The model must match both captions and images correctly, as in \citet{thrush2022winoground}.
% We test spatial relations, temporal relations, verbs and adjectives. As in \dataname, the verb and spatial attributes either have one associated noun or two. However, unlike \dataname, we only evaluate ``left'' and ``right'' spatial relations due to insufficient presence of other spatial relations in the COCO data.
% There are 100 pairs of each type.

\section{Text-only Recovery}
\label{sec:text_only}

\begin{table}[t]
\centering
\resizebox{\columnwidth}{!}{  

            \begin{tabular}{lcc}
            \toprule
              $\mathcal{T}(x)$ & \begin{tabular}{@{}c@{}}Embed. \\ size\end{tabular} & Avg. EM (\%) \\
              \midrule
              CLIP ViT-B/32 & 512 & 13.2 \\
              CLIP ViT-L/14 & 768 & 28.5 \\
              negCLIP ViT-B/32 & 512 & 28.6 \\
              RoBERTaCLIP ViT-B/32 & 512 & 28.9 \\
              \midrule
              SBERT & 768 & \textbf{41.6}\\
              \midrule
              Proof-of-concepT5 & 1024 & \underline{\emph{92.9}} \\
              \bottomrule
            \end{tabular}
        }
\caption{Average EM performance of each text encoder on \dataname, not including multiples and negations (reported in Table \ref{tab: all_negations_multiples}). 
% \jack{Is this the new proof of concept model?} 
}
\label{tab:average_em}
\end{table}

\begin{table*}[h]
\resizebox{\textwidth}{!}{ 
\begin{tabular}{lcc@{}c@{}c@{}cc@{}c@{}c@{}c@{}c@{}c@{}c@{ }c@{}c@{}c@{}c}
 &\textbf{0 attributes} &\multicolumn{4}{c}{\textbf{1 attribute}} &\multicolumn{10}{c}{\textbf{2 attributes}} \\\cmidrule(lr){3-6}\cmidrule(lr){7-16}
& &1 adj. &1 spatial &\multicolumn{1}{c}{\begin{tabular}[c]{@{}c@{}}1 1-obj \\ verb\end{tabular}} &\multicolumn{1}{c}{\begin{tabular}[c]{@{}c@{}}1 2-obj \\ verb\end{tabular}} &2 adj. &\multicolumn{1}{c}{\begin{tabular}[c]{@{}c@{}}1 adj + \\ 1 spatial\end{tabular}} &\multicolumn{1}{c}{\begin{tabular}[c]{@{}c@{}}1 adj + \\ 1 1-obj \\ verb\end{tabular}} &\multicolumn{1}{c}{\begin{tabular}[c]{@{}c@{}}1 adj + \\ 1 2-obj \\ verb\end{tabular}} &\multicolumn{1}{c}{\begin{tabular}[c]{@{}c@{}}1 spatial \\ + 1 1-obj \\ verb\end{tabular}} &\multicolumn{1}{c}{\begin{tabular}[c]{@{}c@{}} 1 spatial \\ + 1 2-obj \\verb\end{tabular}} &2 spatial &\multicolumn{1}{c}{\begin{tabular}[c]{@{}c@{}}1 temp. + \\ 1 1-obj \\ verb\end{tabular}} & \multicolumn{1}{c}{\begin{tabular}[c]{@{}c@{}}1 temp. + \\ 1 2-obj \\ verb\end{tabular}} &2 verbs \\\midrule
\multicolumn{1}{c}{\textbf{\begin{tabular}[c]{@{}c@{}}1 unique \\ object\end{tabular}}} &\multicolumn{1}{l}{\textit{a cat}}& \textit{\begin{tabular}[l]{l}an orange\\ cat\end{tabular}} &\textit{\begin{tabular}[l]{l}a cat on\\ the left\end{tabular}} &\textit{\begin{tabular}[l]{l}a cat \\ yawning\end{tabular}} & - & \textit{\begin{tabular}[l]{l}an orange\\ and spotted\\ cat\end{tabular}} &\textit{\begin{tabular}[l]{l}an orange\\ cat on \\ the left\end{tabular}}& \textit{\begin{tabular}[l]{l}an orange\\ cat \\ yawning\end{tabular}} & - &\textit{\begin{tabular}[l]{l}a cat on \\ the left \\yawning\end{tabular}} & - & - &\textit{\begin{tabular}[l]{l}a cat\\ before \\ yawning\end{tabular}} & - & - \\
CLIP ViT-B/32 &\multicolumn{1}{r}{3.0} &\multicolumn{1}{r}{38.2} &\multicolumn{1}{r}{34.6} &\multicolumn{1}{r}{15.4} & &\multicolumn{1}{r}{14.4} &\multicolumn{1}{r}{47.0} &\multicolumn{1}{r}{36.4} & &\multicolumn{1}{r}{15.6} & & &\multicolumn{1}{r}{17.0} & & \\
CLIP ViT-L/14 &\multicolumn{1}{r}{33.0} &\multicolumn{1}{r}{81.8} &\multicolumn{1}{r}{53.2} &\multicolumn{1}{r}{71.6} & &\multicolumn{1}{r}{23.2} &\multicolumn{1}{r}{43.8} &\multicolumn{1}{r}{62.6} & &\multicolumn{1}{r}{6.2} & & &\multicolumn{1}{r}{58.6} & & \\
negCLIP ViT-B/32 &\multicolumn{1}{r}{2.3} &\multicolumn{1}{r}{42.2} &\multicolumn{1}{r}{57.2} &\multicolumn{1}{r}{20.6} & &\multicolumn{1}{r}{20.8} &\multicolumn{1}{r}{63.6} &\multicolumn{1}{r}{50.4} & &\multicolumn{1}{r}{22.8} & & &\multicolumn{1}{r}{42.4} & & \\
RoBERTa-CLIP ViT-B/32 &\multicolumn{1}{r}{1.3} &\multicolumn{1}{r}{17.0} &\multicolumn{1}{r}{89.4} &\multicolumn{1}{r}{30.0} & &\multicolumn{1}{r}{42.2} &\multicolumn{1}{r}{83.4} &\multicolumn{1}{r}{59.8} & &\multicolumn{1}{r}{5.6} & & &\multicolumn{1}{r}{28.4} & & \\
SBERT &\multicolumn{1}{r}{54.0} &\multicolumn{1}{r}{91.8} &\multicolumn{1}{r}{91.8} &\multicolumn{1}{r}{78.4} & &\multicolumn{1}{r}{35.6} &\multicolumn{1}{r}{85.6} &\multicolumn{1}{r}{76.8} & &\multicolumn{1}{r}{21.2} & & &\multicolumn{1}{r}{57.6} & & \\
\cmidrule{1-16}
\multicolumn{1}{c}{\textbf{\begin{tabular}[c]{@{}c@{}}2 unique \\ objects\end{tabular}}} &\textit{\begin{tabular}[l]{l}a cat and\\ a dog\end{tabular}} & \textit{\begin{tabular}[l]{l}an orange\\ cat and a \\ dog / a cat \\ and a \\ brown dog\end{tabular}} & \textit{\begin{tabular}[l]{l}a cat to\\ the left \\ of a dog\end{tabular}} &\textit{\begin{tabular}[l]{l}a cat \\ yawning \\ and a \\dog\end{tabular}} & \textit{\begin{tabular}[l]{l}a cat\\ chasing \\ a dog\end{tabular}} & \textit{\begin{tabular}[l]{l}an orange\\ cat and a \\ brown dog\end{tabular}} &\textit{\begin{tabular}[l]{l}an orange \\cat to the \\ left of a \\ dog\end{tabular}} & \textit{\begin{tabular}[l]{l}a cat \\yawning \\ and a \\ brown dog\end{tabular}}  &\textit{\begin{tabular}[l]{l}a cat \\chasing \\a brown \\dog / an \\orange cat \\chasing \\a dog\end{tabular}}&\textit{\begin{tabular}[l]{l}a cat \\yawning \\to the \\left of a \\dog\end{tabular}} & \textit{\begin{tabular}[l]{l}a cat \\chasing a \\dog on the \\left\end{tabular}} &\textit{\begin{tabular}[l]{l}a cat on \\the right \\and a dog \\on the left\end{tabular}} &\textit{\begin{tabular}[l]{l}a cat \\before \\yawning \\and a dog\end{tabular}} &\textit{\begin{tabular}[l]{l}a cat \\before \\chasing \\a dog\end{tabular}} &\textit{\begin{tabular}[l]{l}a cat \\yawning \\and a dog \\stretching\end{tabular}} \\
CLIP ViT-B/32 &\multicolumn{1}{r}{13.4} &\multicolumn{1}{r}{15.8}&\multicolumn{1}{r}{6.2} &\multicolumn{1}{r}{1.6} &\multicolumn{1}{r}{6.6} &\multicolumn{1}{r}{17.8} &\multicolumn{1}{r}{6.2} &\multicolumn{1}{r}{0.8} &\multicolumn{1}{r}{7.2} &\multicolumn{1}{r}{8.8} &\multicolumn{1}{r}{4.2} &\multicolumn{1}{r}{1.2} &\multicolumn{1}{r}{1.6} &\multicolumn{1}{r}{3.4} &\multicolumn{1}{r}{1.8} \\
CLIP ViT-L/14 &\multicolumn{1}{r}{48.4} &\multicolumn{1}{r}{30.2}&\multicolumn{1}{r}{17.8} &\multicolumn{1}{r}{5.0} &\multicolumn{1}{r}{37.2} &\multicolumn{1}{r}{23.6} &\multicolumn{1}{r}{14.0} &\multicolumn{1}{r}{1.4} &\multicolumn{1}{r}{21.0} &\multicolumn{1}{r}{16.6} &\multicolumn{1}{r}{7.8} &\multicolumn{1}{r}{0.6} &\multicolumn{1}{r}{0.2} &\multicolumn{1}{r}{21.6} &\multicolumn{1}{r}{4.8} \\
negCLIP ViT-B/32 &\multicolumn{1}{r}{52.8} &\multicolumn{1}{r}{41.4}&\multicolumn{1}{r}{36.2} &\multicolumn{1}{r}{12.2} &\multicolumn{1}{r}{39.4} &\multicolumn{1}{r}{35.6} &\multicolumn{1}{r}{24.8} &\multicolumn{1}{r}{4.8} &\multicolumn{1}{r}{24.2} &\multicolumn{1}{r}{28.6} &\multicolumn{1}{r}{16.4} &\multicolumn{1}{r}{12.4} &\multicolumn{1}{r}{7.4} &\multicolumn{1}{r}{19.2} &\multicolumn{1}{r}{8.8} \\
RoBERTa-CLIP ViT-B/32 &\multicolumn{1}{r}{47.6} &\multicolumn{1}{r}{29.0}&\multicolumn{1}{r}{34.0} &\multicolumn{1}{r}{14.6} &\multicolumn{1}{r}{48.8} &\multicolumn{1}{r}{24.0} &\multicolumn{1}{r}{20.7} &\multicolumn{1}{r}{8.0} &\multicolumn{1}{r}{29.8} &\multicolumn{1}{r}{30.5} &\multicolumn{1}{r}{39.2} &\multicolumn{1}{r}{5.6} &\multicolumn{1}{r}{0.4} &\multicolumn{1}{r}{26.8} &\multicolumn{1}{r}{1.6} \\
SBERT &\multicolumn{1}{r}{56.0} &\multicolumn{1}{r}{41.8}&\multicolumn{1}{r}{44.8} &\multicolumn{1}{r}{11.2} &\multicolumn{1}{r}{48.8} &\multicolumn{1}{r}{32.8} &\multicolumn{1}{r}{30.3} &\multicolumn{1}{r}{8.0} &\multicolumn{1}{r}{30.2} &\multicolumn{1}{r}{31.9} &\multicolumn{1}{r}{23.8} &\multicolumn{1}{r}{3.0} &\multicolumn{1}{r}{4.4} &\multicolumn{1}{r}{4.8} &\multicolumn{1}{r}{10.8} \\
\bottomrule
\end{tabular}}
\caption{Prompt example and exact match (\% EM) score of reconstruction from 
all models, averaged over several hundred instances each. As inputs become more compositional, vision-language text encoders perform increasingly poorly on text reconstruction.
% \kw{When there is only an object, why is the performance worse than with 1obj+ 1 attribute? Also, why 1obj+1 adj+1spatial suddenly get very high performance?  }
% \jack{mention that these are averages over several 100 instances each}
}\label{tab: all_em}
\end{table*}

For a given text encoder $\mathcal{T}$, our first step is to obtain a training corpus of representations to fit a decoding probe $\mathcal{P}(x | \mathcal{T}(x))$. We use (just the text of) Conceptual Captions 3M \cite{sharma2018conceptual} (CC3M) split into a 90/10 train/val set; this corpus consists of cleaned alt-texts from web images, and thus is similar to the pretraining corpora of many VL models. For $\mathcal{P}$, we use T5-large: specifically, we condition the decoder on $\mathcal{T}(x)$, followed by a linear transformation and layer normalization. We train using Adafactor \cite{shazeer2018adafactor} with a batch size of 512 for 4 epochs over CC3M; we select checkpoints with the lowest val loss. Models are trained using 4xA6000 GPUs with 48GB of memory using Transformers \cite{wolf2019huggingface} and accelerate\footnote{\url{https://github.com/huggingface/accelerate}}. % to a text decoder to generate the original caption from each of these representations; 
% 3) we obtain representations from the VL text encoder of the test prompts in our dataset; and
At evaluation time, we generate captions for \dataname set using beam=5 search.%  and evaluate them. 

\paragraph{Text Models.} We evaluate several $\mathcal{T}$ models: CLIP ViT-B/32 (12 layers, 512 dim) and ViT-L/14 (12 layers, 768 dim) \cite{radford2021learning}, % BLIP \cite{li2022blip},
CLIP with a RoBERTa-pretrained text encoder \cite{Liu2019RoBERTaAR, ilharco_gabriel_2021_5143773}, and \citet{yuksekgonul2022and}'s more-order-aware CLIP encoder finetuned with hard negatives, negCLIP. For comparison, we also consider the uni-modal SentenceBERT \cite{reimers2019sentence} model \texttt{all-mpnet-base-v2}, which is trained on several sentence similarity datasets including COCO captions \cite{lin2014microsoft}.

\paragraph{Proof-of-concepT5} We also consider a T5-large text encoder that produces a single vector output via mean pooling over the token embeddings. In contrast to the other fixed encoders, we fine-tune this model on CC3M, like an autoencoder\footnote{There is no overlap between CC3M and \dataname.}. Then, we use the resulting encoder as a feature extractor, and hand a dimension-shuffled version of the resulting embeddings to the probe. 
% \jack{There is no overlap between the two datasets, right?} 
This ``proof of concept'' encoder is specifically optimized to generate a vector from which a T5 model can decode the full sentence, and serves to validate that our probe setup is even possible.

\paragraph{Evaluation.} We evaluate using exact match (EM). While we report BLEU scores in the Appendix, % testing whether the right keywords appear in a valid order (ignoring extra words) are in the appendix; however, these two methods do not capture the bottleneck nature of the text encoder. For example, predicting
for our high-precision setting, partial credit metrics are too generous, e.g., generating ``a reporter on top of a penguin'' as ``a penguin on top of a hill'' scores 48 BLEU-4 points. Similarly for BERT-Score \cite{bert-score}, where generating ``two rabbits and three shrimps'' as ``four of the shrimps and a rabbit'' scores 0.91 F1.

% \kw{B-4 ?}
% but almost all of the information is lost. On the flip side, predicting ``a chimpanzee'' as ``chimpanzee crawling a tree'' scores full points on the keyword evaluation — although false information is added, which could cause problems downstream.
% \amita{We need to discuss evaluation of \evalname as well (image and text scores per the description in the Winoground paper) -- think of where to put this.}

\subsection{Text-only Recovery Results}
\label{sec:compprompts_results}

Table~\ref{tab:average_em} presents the average exact match of each model over the corpus of prompts in \dataname, excluding negations and multiples, which are reported in Table \ref{tab: all_negations_multiples}. The proof-of-concepT5 model's high performance illustrates that it is possible in theory to nearly exactly decode all captions in \dataname using T5-large, given the ``right'' encoding\footnote{Most errors made by proof-of-concepT5 are minor e.g., ``two physicians on the right'' $\rightarrow$ ``two physician on the right''.}. Beyond proof-of-concepT5, the best performing model is SBERT; and the best performing multimodal model is RoBERTa-CLIP.

\subsection{Fine-Grained Results on Different Prompt Types}

Table~\ref{tab: all_em} contains EM results of all models
% , the better-performing original CLIP model, 
on the various types of prompts in \dataname. % as well as the EM of CLIP ViT-B/32 on each. %The average EM of each model is shown in Table \ref{tab:average_em}, with detailed results in the appendix.
% From these results, it's clear to see that there is a significant gap between what is currently encoded in a single vector by VL models' text encoders, and what could be, as shown by the SBERT results.  --- this is touched on in proof of concept
A separate study on multiples and negations in Table \ref{tab: all_negations_multiples} shows that text encoders struggle to encode those as well. These results show that it is fairly difficult to decode input sentences from text representations for most VL models, with increasingly compositional categories proving more difficult (e.g., ``an orange cat'' to ``an orange cat yawning'' to ``an orange cat chasing a dog'').

% \subsection{Model performance on various facets of language}

%Models clearly begin to suffer when the prompts become more compositional (``an orange cat'' to ``an orange cat yawning'' to ``an orange cat chasing a dog''). They perform poorly on average on relationships between objects, whether the relationship is spatial, temporal, or otherwise.

% Below, we discuss the various aspects of language evaluated by our dataset, with observations of the performance of various models on the same. 

\paragraph{Spatial relations.} Text encoders of VL models struggle to represent spatial relations (average 23.7 EM), particularly those between two objects (average 13.8 EM). SBERT, in comparison, scores 36.9 and 22.3 EM, respectively. 

\paragraph{Temporal relations.} VL models perform poorly on temporal relations, scoring on average 17.1 EM. In comparison, SBERT scores 29.6 EM --- likely because temporal relations appear more frequently in language than in web alt-text. %RoBERTa-CLIP does surprisingly poorly despite its language pre-training (11.2 EM points).

\paragraph{Transitive vs intransitive verbs and prepositions.} On transitive verbs (e.g., ``chasing''), CLIP ViT-B/32 and ViT-L/14 do worse by an average of 21 EM than vs. intransitive verbs (e.g., ``yawning''), whereas negCLIP and RoBERTa-CLIP do better by an averaged 18.7 points. On transitive prepositions (``to the left of'') instead of intransitive (``on the left''), all models do worse by an averaged 35 EM. 

\paragraph{Negations and multiples.} Models perform poorly on negations (average EM 13.0) and multiples (average EM 5.1). This agrees with previous observations that VL models struggle with counting \cite{segui2015learning,parcalabescu2020seeing}. 

\paragraph{Prompts where word order matters.} VL text encoders struggle to capture word order: on prompts where word order matters less (e.g., ``a cat and a dog''), they score an average of 34 EM, but where word order matters more, they score an average of 15.8 EM. The failure cases are often caused by assigning attributes to nouns incorrectly, as highlighted in the Appendix. This extends \citet{thrush2022winoground}'s and \citet{yuksekgonul2022and}'s finding that contrastive VL models can behave like bags-of-words --- this issue manifests just in the text encoder as well.

\paragraph{Adjectives and verbs.} VL models perform relatively well in the basic one-object, one-attribute setting on both adjectives (average EM 44.8) and verbs (average EM 34.5): even higher than the zero-attribute setting, where error analysis reveals they tend to hallucinate information (``a tarantula'' $\rightarrow$ ``a tarantula in a hand''). While these numbers are well behind SBERT (EM 91.8 and 78.4 respectively), they agree with previous observations that VL models exhibit good visual recognition of basic adjectives and actions \cite{radford2021learning}.

\paragraph{Compositionality.} Text encoders struggle with increasingly compositional information, e.g., the probe decodes SBERT(``a dentist after examining an ape'') $\rightarrow$ ``an ape after examining a dentist''. On average, performance on two unique objects drops by 49\% from their performance on one unique object (for CLIP ViT-B/32, it drops 71\%). VL model performance drops on average by 35\% when the prompt contains two attributes compared to one.

\begin{table}[t]
\resizebox{\columnwidth}{!}{ 
\begin{tabular}{lccccc}
 &\textbf{0 attributes} &\multicolumn{4}{c}{\textbf{1 attribute}}\\\cmidrule(lr){3-6}
& &1 adj. &1 spatial &\multicolumn{1}{c}{\begin{tabular}[c]{@{}c@{}}1 1-obj \\ verb\end{tabular}} &\multicolumn{1}{c}{\begin{tabular}[c]{@{}c@{}}1 2-obj \\ verb\end{tabular}} \\\midrule
\multicolumn{1}{c}{\textbf{\begin{tabular}[c]{@{}c@{}}1 unique\\ object +\\multiples\end{tabular}}} &\multicolumn{1}{l}{\textit{two cats}} &\textit{\begin{tabular}[l]{l}two orange\\ cats\end{tabular}} & \textit{\begin{tabular}[l]{l}two cats on\\ the left\end{tabular}} &\textit{\begin{tabular}[l]{l}two cats\\ yawning\end{tabular}} &- \\
CLIP ViT-B/32 &\multicolumn{1}{r}{0.3} &\multicolumn{1}{r}{15.0} &\multicolumn{1}{r}{9.6} &\multicolumn{1}{r}{9.4} & \\ 
CLIP ViT-L/14 &\multicolumn{1}{r}{3.0} &\multicolumn{1}{r}{14.8} &\multicolumn{1}{r}{10.8} &\multicolumn{1}{r}{16.8} & \\
negCLIP ViT-B/32 &\multicolumn{1}{r}{1.0} &\multicolumn{1}{r}{9.4} &\multicolumn{1}{r}{11.2} &\multicolumn{1}{r}{12.8} & \\
RoBERTa CLIP ViT-B/32 &\multicolumn{1}{r}{0.7} &\multicolumn{1}{r}{4.6} &\multicolumn{1}{r}{16.0} &\multicolumn{1}{r}{6.4} & \\
SBERT &\multicolumn{1}{r}{37.3} &\multicolumn{1}{r}{56.6} &\multicolumn{1}{r}{48.4} &\multicolumn{1}{r}{33.2} & \\
\cmidrule{1-6}
\multicolumn{1}{c}{\textbf{\begin{tabular}[c]{@{}c@{}}2 unique\\ objects +\\multiples\end{tabular}}} &\textit{\begin{tabular}[c]{@{}l@{}}two cats \\and \\four dogs\end{tabular}} &- & \textit{\begin{tabular}[l]{l}two cats to\\ the left of \\four dogs\end{tabular}} &- &\textit{\begin{tabular}[l]{l}two cats \\ chasing \\four dogs\end{tabular}} \\
CLIP ViT-B/32 &\multicolumn{1}{r}{0.0} & &\multicolumn{1}{r}{0.0} & &\multicolumn{1}{r}{0.0} \\
CLIP ViT-L/14 &\multicolumn{1}{r}{0.0} & &\multicolumn{1}{r}{0.0} & &\multicolumn{1}{r}{0.0} \\
negCLIP ViT-B/32 &\multicolumn{1}{r}{0.0} & &\multicolumn{1}{r}{0.0} & &\multicolumn{1}{r}{0.2} \\
RoBERTa CLIP ViT-B/32 &\multicolumn{1}{r}{0.6} & &\multicolumn{1}{r}{0.2} & &\multicolumn{1}{r}{0.6} \\
SBERT &\multicolumn{1}{r}{0.0} & &\multicolumn{1}{r}{0.0} & &\multicolumn{1}{r}{0.0} \\
\midrule
\multicolumn{1}{c}{\textbf{\begin{tabular}[c]{@{}c@{}}1 unique\\ object +\\negation\end{tabular}}} &- & \textit{\begin{tabular}[l]{l}a cat that\\ is not \\orange\end{tabular}} &\textit{\begin{tabular}[l]{l}a cat that\\ is not on \\the left\end{tabular}}& \textit{\begin{tabular}[l]{l}a cat that\\ is not \\yawning\end{tabular}} &- \\
CLIP ViT-B/32 & &\multicolumn{1}{r}{18.0} &\multicolumn{1}{r}{7.4} &\multicolumn{1}{r}{6.8} & \\
CLIP ViT-L/14 & &\multicolumn{1}{r}{10.4} &\multicolumn{1}{r}{6.4} &\multicolumn{1}{r}{7.8} & \\
negCLIP ViT-B/32 & &\multicolumn{1}{r}{1.8} &\multicolumn{1}{r}{2.2} &\multicolumn{1}{r}{14.2} & \\
RoBERTa CLIP ViT-B/32 & &\multicolumn{1}{r}{43.8} &\multicolumn{1}{r}{50.4} &\multicolumn{1}{r}{58.4} & \\
SBERT & &\multicolumn{1}{r}{32.8} &\multicolumn{1}{r}{4.6} &\multicolumn{1}{r}{19.4} & \\
\cmidrule{1-6}
\multicolumn{1}{c}{\textbf{\begin{tabular}[c]{@{}c@{}}2 unique\\ objects +\\negation\end{tabular}}} &- &- & \textit{\begin{tabular}[l]{l}a cat that \\is not to \\the left \\of a dog\end{tabular}} &- &\textit{\begin{tabular}[l]{l}a cat that \\is not \\chasing \\a dog\end{tabular}} \\
CLIP ViT-B/32 & & &\multicolumn{1}{r}{0.6} & &\multicolumn{1}{r}{0.8} \\
CLIP ViT-L/14 & & &\multicolumn{1}{r}{0.4} & &\multicolumn{1}{r}{0.4} \\
negCLIP ViT-B/32 & & &\multicolumn{1}{r}{2.6} & &\multicolumn{1}{r}{1.8} \\
RoBERTa CLIP ViT-B/32 & & &\multicolumn{1}{r}{12.6} & &\multicolumn{1}{r}{13.6} \\
SBERT & & &\multicolumn{1}{r}{2.6} & &\multicolumn{1}{r}{2.6} \\
\bottomrule
\end{tabular}}
\caption{All models' EM on prompts that contain multiples or negations. Text recovery of these inputs is very poor, likely because multiples and negations tend to be infrequent in image captions.}\label{tab: all_negations_multiples}
\end{table}

%\subsection{Impact of model design on performance}
%We now discuss the impact of various types of modifications to CLIP captured by the models we selected to evaluate, by looking at the differences between, e.g. negCLIP and CLIP: 
\subsection{Fine-Grained Results for Different Model Designs}
\label{sec:compprompts_model_design}

\paragraph{Pre-training the text encoder helps, especially on negations.} The average EM of RoBERTa-CLIP on prompts without multiples or negations is 15.7 points higher than CLIP ViT-B/32. However, on the prompts that do include negations, its average EM is 29 points higher. %On multiples it is about 1 point lower.
This provides evidence that text pre-training the text encoder helps negations, presumably because negations are less likely in alt-texts compared to other settings. 

\paragraph{Increasing model size helps overall, but not on spatial relations.} The average EM of CLIP ViT-L/14 on prompts that do not include spatial relations is 20.7 points higher than CLIP ViT-B/32. However, on the prompts that do include spatial relations, its average EM is only 4 points higher. The modest increase of text encoder size in the CLIP training regime appear less reliable for encoding spatial relations than text pre-training or hard negatives (though, more significant scaling could be beneficial, as in Imagen \cite{saharia2022photorealistic}).

\paragraph{Hard negatives from \citet{yuksekgonul2022and} help, especially where word order matters.} % Training CLIP with hard negatives targeting word order shuffling helps model performance significantly ---
On average, negCLIP does 15.4 points better than CLIP. On prompts where word order matters (e.g. ``a cat chasing a dog''), it scores 16.3 points higher; on prompts where word order does not matter (e.g. ``a cat and a dog''), it scores 12.8 points higher. 

% already addressed with proof of concept
% \paragraph{More information \textit{can} be squeezed into one vector.} The clearly superior results from SBERT at the same embedding size as CLIP ViT-L/14 show that more information can indeed be packed into one vector. 
% \textcolor{blue}{
\subsection{Incorrect Model Predictions}
We manually inspect models' incorrect predictions. Decoded VL text encoder predictions 
% \jack{which? CLIP?} \amita{All of them, actually -- this particular example is a common pattern shown by CLIP B/32, CLIP L/14 and negCLIP.} 
often come close (e.g. ``three shrimps'' $\rightarrow$ ``three of shrimp'' is a pattern shown by CLIP ViT-B/32, CLIP ViT-L/14 and negCLIP), whereas SBERT's incorrect decodings fall further afield (e.g. ``three gardeners'' $\rightarrow$ ``three gardeners and a third man.''). Thus, while the superior results of the unimodal SBERT compared to the VL text encoders when evaluated in the same frozen-encoder setting (including CLIP ViT-L/14, which has the same text embedding size) show that there is significant room for improvement for VL text encoders, the types of errors made by each model may not be fully captured by EM. Nonetheless, % the gap is potentially slightly smaller than EM makes it seem. However,
EM remains an % the most 
appropriate metric for our high-precision setting, as discussed in Section \ref{sec:text_only}.
% }

% \input{tables/example_table.tex}

% \input{table2.tex}

\section{Experiments and Results in the Multi-modal Setting}
\begin{figure}
    \centering
    \includegraphics[width=0.9\columnwidth]{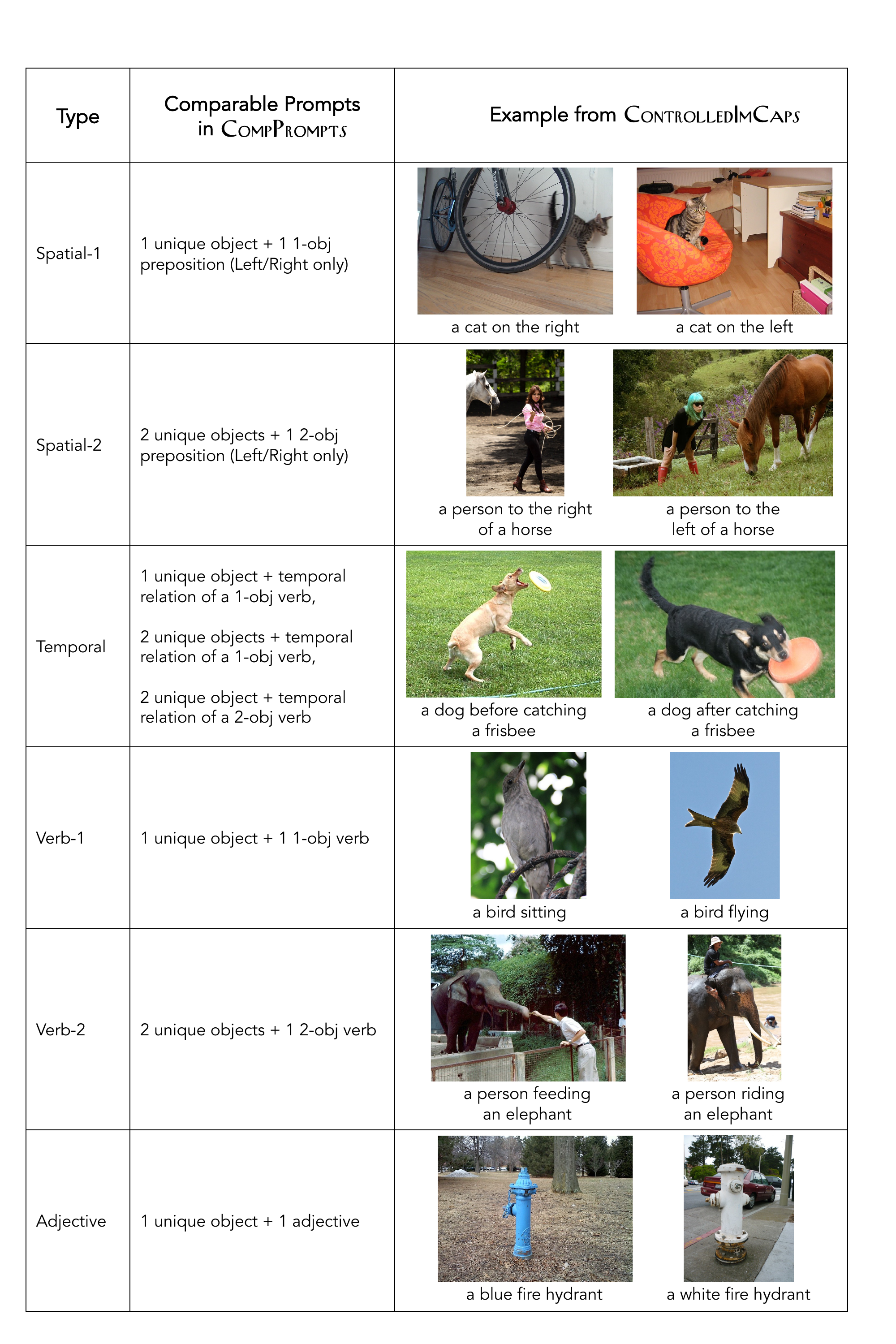}
    \caption{Each attribute in \evalname, with comparable prompts in \dataname and an example. 
    % \jack{This figure is bumping into the margin. I don't dare try to fix it on the train, but maybe we can throw it in a resizebox?}
    % We also show in a separate evaluation that this behavior continues downstream.
    }
    \label{fig:cic_examples}
\end{figure}

We investigate the hypothesis that if a textual property cannot be decoded
from the VL text encoder's vector representation with a highly expressive decoder (like T5-Large), then it also cannot be readily
modeled in the multimodal setting. % JMH: I don't think we need this summary here. In Section \ref{sec:compprompts_results}, we discuss the performance of VL text encoders on various textual properties. Here, we evaluate the VL model as a whole on image pairs targeting these textual properties.
\evalname studies the attributes from \dataname in the multimodal setting. We then compare the text encoder performance of a VL model on a particular prompt type in \dataname with the performance of the overall VL model on that prompt type in \evalname. As discussed in Section \ref{sec:evalname}, the two captions in every example differ by only one word which changes or inverts the relation, allowing us to perform fine-grained analyses in controlled settings without conflating multiple types of compositional reasoning. Figure \ref{fig:cic_examples} depicts the six types of attributes studied in \evalname, their corresponding prompt type in \dataname, and an example of each.

\paragraph{VL Models.} We evaluate the same VL models as in Section \ref{sec:text_only}: CLIP ViT-B/32, CLIP ViT-L/14, CLIP with a RoBERTa-pretrained text encoder \cite{Liu2019RoBERTaAR,ilharco_gabriel_2021_5143773}, and negCLIP \cite{yuksekgonul2022and}. Each of these models can return a score when given an image and a caption, representing how well they match.

\paragraph{Evaluation.} We follow the evaluation scheme from Winoground \cite{thrush2022winoground}: for a given pair of images with corresponding captions, we measure both a \textit{text score}, the fraction of instances where a model scores the correct caption higher than the incorrect caption when given an image, and an \textit{image score}, the fraction of instances where a model scores the correct image higher than the incorrect image when given a caption.

% \jack{We are missing a description of \evalname's setup and metrics and examples. Feels like there could be a whole paragraph here and a figure with examples.}
\subsection{Multi-modal Results}
Table \ref{tab:downstream_vitl} presents the results of CLIP ViT-L/14 on both \dataname and \evalname (all model results in Appendix). The \dataname results correspond to the prompt type(s) most closely matching the captions in \evalname (specified in Figure \ref{fig:cic_examples}). For the spatial relations, for this table alone, we calculate the EM on the data points in \dataname containing ``left'' and ``right'' spatial relations only due to lack of sufficient support in COCO for other spatial relations, as discussed in Section \ref{sec:evalname}. On prompt types where the text encoder performance on \dataname is poor, the overall model performance on \evalname is also poor: showing that the text encoder does indeed bottleneck VL models' compositionality. 
% \jack{I think we should try to replace "Evidently" with the concrete evidence once we figure the right metric :-)}

\begin{table}[t]
\centering
\resizebox{0.9\columnwidth}{!}{  

            \begin{tabular}{lccc}
            \toprule
              Prompt Type & \begin{tabular}{@{}c@{}}EM on \\ \dataname\end{tabular} & \begin{tabular}{@{}c@{}}\funfont{CIC}\\Image \\ score\end{tabular} & \begin{tabular}{@{}c@{}}\funfont{CIC}\\Text \\ score\end{tabular} \\
              \midrule
              Spatial 1-obj L/R & 28.5 & 4.0 & 15.0 \\
              Spatial 2-obj L/R & 4.4 & 4.0 & 8.0 \\
              Temporal & 26.8 & 7.0 & 30.0 \\
              Verb 1-obj & 71.6 & 84.0 & 89.0\\
              Verb 2-obj & 37.2 & 46.0 & 63.0\\
              Adjectives & 81.8 & 65.0 & 85.0\\
              \bottomrule
            \end{tabular}
        }
\caption{Performance of CLIP ViT-L/14 text encoder (\%) on the equivalent prompts in \dataname, and performance of CLIP ViT-L/14 full model on \evalname (\funfont{CIC}). On the types of prompts where the text encoder performs poorly, so too does the overall model. 
% \jack{can we fit \evalname into the headers for the second two columns?} 
% \jack{what are image/text score? same setup as winoground?}
}
\label{tab:downstream_vitl}
\end{table}

We see similar findings per prompt type and model design as those discussed in Section \ref{sec:compprompts_model_design}.

\subsection{Fine-Grained Results on Different Prompt Types}
We discuss findings on the prompt types in \evalname, with 95\% confidence intervals.
\paragraph{Models do poorly on spatial relations.} On average, VL models perform poorly on spatial relations, achieving an average image $|$ text score of 2.5 $|$ 12.4 ($\pm$ 2.2 $|$ 3.7). Their text encoder performance on the corresponding prompts in \dataname was similarly poor, with an average EM of 27.8. This agrees with \citet{kamath2023whatsup}, which shows that VL models struggle with spatial relations.

\paragraph{Models do poorly on temporal relations.} VL performs poorly on temporal relations, with an average image $|$ text score of 5.3 $|$ 30.8 ($\pm$ 2.7 $|$ 4.8). Their text encoder performance on \dataname temporal reasoning was similarly low at 18.9 EM.  

\paragraph{Models do well on verbs and adjectives.} VL models perform well on verbs (average image $|$ text score 65.4 $|$ 78.1, $\pm$ 5.0 $|$ 4.8) and even better on adjectives (average image $|$ text score 78.5 $|$ 89.0, $\pm$ 7.0 $|$ 3.5), mirroring their text encoder performance on \dataname, where the average EM for verbs and adjectives were 33.7 and 44.8 respectively.

\paragraph{Two-object verbs are more difficult than one-object verbs.} We find that for all models, two-object verbs are harder than one-object verbs, with the former achieving an image $|$ text score of 52.3 $|$ 68.5 and the latter 78.5 $|$ 87.8 (with $p < 0.05$ under the Wilcoxon signed-rank test). This follows performance on \dataname for ViT-B/32 and ViT-L/14, but not for negCLIP and RoBERTa-CLIP, hinting that ability to reconstruct is necessary but not sufficient, as discussed in Section \ref{sec:necessary_insufficient}. 
% This could imply that the ability to recover the text input is a necessary but insufficient condition for overall model performance.

\subsection{Fine-grained results on different model
design choices}
We discuss findings on the model designs in \evalname. All findings are statistically significant at $p<0.05$ using the Wilcoxon signed-rank test to compare models.
% \amita{Need to calculate p-values here and see if it even makes sense to discuss these results}
\paragraph{Pre-training the text encoder improves text score on verbs.} RoBERTa-CLIP obtains a higher text score than CLIP ViT-B/32 (78.0 vs 68.0), as well as a higher EM on the prompts in \dataname corresponding to verbs (39.4 vs 11.0).
% RoBERTa CLIP ViT-B/32 obtains a higher image | text score than CLIP ViT-B/32 (35.0 | 49.3 versus 32.8 | 47.2) as well as a higher EM on the prompts in \dataname corresponding to the inputs in \evalname (39.2 versus 15.9). On verbs, this is even more evident, with RoBERTa-CLIP scoring an image | text score of 62.5 | 78.0 and a \dataname EM of 39.4, and with CLIP ViT-B/32 scoring an image | text score of 57.0 | 68.0 and a \dataname EM of 11.0. 
% \amita{Why could this be?}

\paragraph{Increasing model size does not help on spatial or temporal reasoning.} On both spatial and temporal reasoning inputs, ViT-L/14 performance on \evalname was not statistically significantly higher than that of ViT-B/32. 
% ViT-L/14 scores higher than ViT-B/32 on both image | text (65.0 | 79.0 versus 62.7 | 76.0) and \dataname EM (63.5 vs 20.1). However, on spatial and temporal relations, this is not the case: ViT-L/14 scores an average of 5.0 | 17.7 image | text and 19.9 EM on \dataname, where ViT-B/32 scores 3.0 | 18.3 and 11.7 EM respectively.

\paragraph{Hard negatives from \citet{yuksekgonul2022and} help where word order matters.} On prompts where word order matters, negCLIP scores an image $|$ text score of 36.5 $|$ 50.0 and a \dataname EM of 27.2, and other models score an average image $|$ text score of 24.8 $|$ 37.5 and a \dataname EM of 21.0. negCLIP also outperforms ViT-B/32 on all prompts on average. 

% \textcolor{blue}{
\subsection{Text reconstruction appears to be necessary...}
\label{sec:necessary}
To study the relationship between text reconstruction and overall model performance beyond Table \ref{tab:downstream_vitl}, we evaluate text reconstruction on \evalname. Specifically, we use the trained T5 decoders from Section \ref{sec:text_only} and try to reconstruct the input when \evalname text inputs are evaluated. On the cases where the reconstruction is \textit{incorrect} according to human evaluation\footnote{For the simple inputs of \evalname, we found human evaluation by the authors tractable, with the added advantage of not penalizing minor errors as EM does.} on either of the two text inputs, the overall model Image Score on \evalname for CLIP ViT-L/14 is zero 96\% of the time, and the Text Score is zero 83\% of the time. This text reconstruction vs. multimodal matching correlation is more direct compared to %relationship between text reconstruction and overall model performance
the similar correlation reported in Table \ref{tab:downstream_vitl} because we compare on the same instances. %but is restricted in scale and scope compared to evaluating text reconstruction on \dataname, which is useful as a broad-spectrum diagnostic tool to determine model weaknesses
% }

% \textcolor{blue}{
\subsection{... but insufficient.} 
\label{sec:necessary_insufficient}
% However, just because the probe can accurately decode a caption from a given text encoder does not mean the multimodal model will be able to
Conversely, just because a model performs well on the \dataname probe does not mean it will perform well on \evalname. %there are some cases where the performance of a VL text encoder on a particular prompt type in \dataname is acceptable, but the performance of the VL model on the corresponding prompt type in \evalname is poor.
For example, ViT-L/14 outperformed ViT-B/32 overall on \dataname, but not (statistically significantly) on \evalname. Also, RoBERTa-CLIP outperforms ViT-B/32 on temporal relations on \dataname, but achieves a worse text score on \evalname. 
% e.g., RoBERTa-CLIP scores 92.3 EM on \dataname prompts targeting spatial relations with 1 object, but on \evalname gets an image | text score of 1.0 | 10.0 on the same prompt type. 
When we evaluate text reconstruction on \evalname, on cases where the reconstruction is \textit{correct} for both text inputs, the overall model Image Score on \evalname for ViT-L/14 is zero 59\% of the time, and the Text Score is zero 47\% of the time. 
This suggests that text recoverability is a necessary but insufficient condition for overall model performance.
The insufficiency is intuitive, as multimodal errors could potentially stem from the image encoder.
% }

\subsection{A Note on Winoground}
We evaluate our four VL models on the Winoground dataset \cite{thrush2022winoground}. They perform poorly, with an average image $|$ text score of 10.3 $|$ 30.8, where random chance is 25.0 $|$ 25.0.
% \jack{compared to what?} 
However, %this number masks a more nuanced finding:
on shorter inputs (5 words or less) which exhibit fewer compositional concepts on average, e.g., ``a bird eats a snake'' $|$ ``a snake eats a bird'', the four models achieve higher scores of 20.4 $|$ 47.2 on average. On longer (over 10 words), more compositional inputs, e.g. ``in the stadium, the person wearing gray outperformed the one wearing blue'' $|$ ``in the stadium, the person wearing blue outperformed the one wearing gray'', models achieve a much lower score of 3.4 $|$ 18.5. This mirrors our finding on \dataname that VL text encoders struggle with increasingly compositional inputs.
% On the non-55, average is 9.8 | 29.3. 

\section{Related work}
Building models capable of reasoning jointly about visual and textual inputs is a long-standing goal of AI \cite{winograd1971procedures}, with potential applications in the fields of vision-language navigation \cite{anderson2018vision}, human-robot interaction \cite{matuszek2012joint}, accessible image captioning \cite{gurari2020captioning}, etc.

Recent challenge datasets have been designed to probe the capacity of multimodal models to represent descriptions of precise visual compositions \cite{johnson2017clevr,suhr2018corpus,hudson2019gqa,thrush2022winoground}. \citet{yuksekgonul2022and} and \citet{yamada2022lemons} study CLIP specifically, demonstrating its shortcomings (and some potential fixes) in terms of modeling syntax. \citet{Ma2022CREPECV} study OpenCLIP models for various types of compositional reasoning, with programmatically sourced hard negatives. Different from these works, our textual probe does not require access to images. 

Our image-and-text evaluation most closely resembles \citet{thrush2022winoground}. However, we stratify the examples based on type of input (e.g., temporal relations) to provide more detailed insights. We also keep our prompts relatively simple, never having more than two objects or two attributes in the input. We believe this is a more realistic goal for our current vision-language models. The word order shuffling aspect is also discussed in \citet{yuksekgonul2022and}. However, as their proposed benchmark does not provide pairs of images with corresponding captions, it is possible to achieve state-of-the-art with a text-only model (specifically, 2-shot ChatGPT\footnote{\url{https://platform.openai.com/docs/api-reference/chat}, using the \texttt{gpt-3.5-turbo} model} \cite{Ouyang2022TrainingLM}, details in Appendix and the recent \citet{hsieh2023sugarcrepe}). While this does not detract from their finding that vision-language models ignore word order, our benchmarks have an additional advantage of being insensitive to text-only priors. % from training not helping on either \dataname (which randomly matches objects, resulting in unusual inputs like ``a crab lifting a rhino'') or \evalname (which has image pairs). 

% referring expressions seem related..
%\cite{mitchell2010natural}

\section{Conclusion and Discussion} 
We present probing results that suggest significant information loss upon text encoding of compositional inputs in vision and language models. This information loss is quantified using \dataname, a test set of increasingly compositional image descriptions, and \evalname, a test set that we use to verify that this information loss affects the performance of multimodal models on compositional inputs.
% Poor text encoder performance correlates with poor performance of the overall multimodal model, showing that this information loss does impact VL model performance.
% information loss affects multimodal models overall \jack{verify the results at \jack{multimodal?} the overall model level. %.re captured by the text encoders of popular constrastive VL models.
Harder negatives, more text pre-training, and larger models all improve encoder quality, but information is still lost even for the most performant models, compared to the uni-modal SBERT as well as a T5-based auto-encoder.

% more research is needed to improve them further.
%It is clear that there is significant headroom for improvement. and that better text encoders are necessary for better performing VL models.
%The various models we tested, which improve on CLIP ViT-B/32 in different ways, have different benefits and drawbacks on model performance as described in the paper, bu

Going forward, even more difficult test sets than \dataname and \evalname might be required to analyze and evaluate vision-language model capabilities. 
% \jack{What about CREPE?} 
Returning to \citet{marktweet}'s tweet from the intro, \emph{``A penguin on Mars wearing a spacesuit and walking a robot dog next to Santa Claus.''}, even our highly accurate proof-of-concepT5 model struggles, predicting: \emph{``compulsory penguin onexposition wearing a spacesuit and walking a dog robot next tohoc''}. To support imaginative text-to-image generation queries (for images that may not exist yet), future work would be well-suited to design text encoders that can generalize to captions  %even more imaginative captions
that contain compositions never-before-seen in web alt-text corpora.

Our probing results suggest two future modeling directions:
% \textcolor{blue}{
(1) Modifying contrastive VL models' training objectives to additionally encourage ability to reconstruct the text input, either through an additional reconstruction loss on the text encoder during finetuning, or through the addition of even harder negatives than \citet{yuksekgonul2022and} and \citet{Ma2022CREPECV}, would be an exciting avenue for future work. 
Alternatives to contrastive training, such as captioning, have also shown promise in recent work \cite{tschannen2023image};
% \jack{What about image captions are scalable image text foundation models?}
% }
% \textcolor{blue}{
and (2) explicitly encouraging linear recovery with a modified loss function: %Our work also raises the question: is there a limit on how much information can be encoded in a single vector in a manner that is linearly recoverable?
while the gap between SBERT and the VL Text encoders can be partially explained by the superior pooling method and training data, SBERT's training objective does not require linear recoverability (whereas CLIP's dot product interaction term might): explicitly encouraging linear text-text recoverability might improve multimodal performance. % also does not need to encode information such that it can be extracted through a simple dot product. 
% }
% We need better text encoders
% Hard negatives are important
% Better pooling methods (a la SBERT) are important
Finally, we hope that \evalname can facilitate research beyond single-vector bottleneck VL models. % allowing for fine-grained evaluation of VL models on specific relations.  

% \jack{Something about how \evalname could be helpful for not just single-vector bottleneck models?}

\section*{Limitations}
First, our probing method involves a pre-trained T5 decoder. It is possible that language biases from the pre-training emerge while decoding from the VL text embedding, e.g., predicting ``a dog chasing a cat'' instead of ``a cat chasing a dog'' because the former is more likely under the T5 decoder's priors from pre-training. However, as the methodology is the same across all models we evaluate, we believe that the evaluation is fair. 
Second, we evaluate with only one probe, whereas probing with complementary methods (e.g., especially deterministic ones, like a convex linear probe) could reveal more insights. 
% Third, while being text-only is an advantage in terms of scale, control and diversity, text encoders that do well on our evaluation may perform worse at object detection if they front compositional relationships more than ``object-ness." 
Third, text encoders that do well on our evaluation may not perform well if directly plugged into a contrastive VL model like CLIP, if the text encoders were not trained to encode the information in a manner that is linearly recoverable.
% Finally, we only consider the English language.

%\subsection{References}

\section*{Acknowledgements}
The authors thank John Hewitt,  Akhila Yerukola, and anonymous reviewers for useful discussion and feedback. 
This work was funded by the Allen Institute for AI. AK was additionally supported by the UCLA Computer Science Department First-Year Fellowship. KC was supported in part by U.S. DARPA MCS Program under contract number N660011924032, U.S. DARPA ECOLE Program No. HR00112390060, and ONR N00014-23-1-2780, and a Sloan Fellowship. The views and conclusions contained herein are those of the authors and should not be interpreted as necessarily representing the official policies, either expressed or implied, of DARPA, or the U.S. Government. 
% The U.S. Government is authorized to reproduce and distribute reprints for governmental purposes notwithstanding any copyright annotation therein.

\bibliography{refs}
\bibliographystyle{acl_natbib}

\clearpage

\appendix

\section{Additional Results}
\label{sec:appendix}
Table \ref{tab:average_bleu} contains average BLEU-4 scores of the models. Table \ref{tab:shuffled} contains a study of model performance on object-attribute association in \dataname. Table \ref{tab:downstream_vitb}, Table \ref{tab:downstream_negclip} and Table \ref{tab:downstream_robertaclip} contain results of other models on \evalname in comparison to \dataname (ViT-L/14 is discussed in Table \ref{tab:downstream_vitl}). Table \ref{tab:aro_text_only} discusses text-only results on the ARO benchmark \cite{yuksekgonul2022and}. 
\begin{table}[h]
\centering
\resizebox{.9\columnwidth}{!}{  

            \begin{tabular}{lcc}
            \toprule
              $\mathcal{T}(x)$ & \begin{tabular}{@{}c@{}}Embed. \\ size\end{tabular} & Avg. BLEU-4 \\
              \midrule
              CLIP ViT-B/32 & 512 & 33.3 \\
              CLIP ViT-L/14 & 768 & 46.0 \\
              negCLIP ViT-B/32 & 512 & 50.2 \\
              RoBERTaCLIP ViT-B/32 & 512 & 52.0 \\
              \midrule
              SBERT & 768 & \textbf{56.7}\\
              \bottomrule
            \end{tabular}
        }
\caption{Average BLEU-4 performance of each text encoder on \dataname, not including multiples and negations. The trend correlates with EM \%, but the evaluation itself is too lenient for our purposes, as described in the main text.}
\label{tab:average_bleu}
\end{table}
\begin{table}[h]
\centering
\resizebox{.9\columnwidth}{!}{  

            \begin{tabular}{lcc}
            \toprule
              $\mathcal{T}(x)$ & \begin{tabular}{@{}c@{}}Embed. \\ size\end{tabular} & Shuffled $\%$ ($\downarrow$)\\
              \midrule
              CLIP ViT-B/32 & 512 & 51.8 \\
              CLIP ViT-L/14 & 768 & 55.5 \\
              negCLIP ViT-B/32 & 512 & \textbf{37.2} \\
              RoBERTaCLIP ViT-B/32 & 512 & 62.8 \\
              \midrule
              SBERT & 768 & 44.2\\
              \bottomrule
            \end{tabular}
        }
\caption{Of the times the model gets the words in the prediction correct, Shuffled $\%$ is the percentage of when it gets the word order incorrect (in the prompts where word order matters, unlike ``cat and dog'' --- specifically, where attributes must be associated with the correct object). Clearly, negCLIP having been trained with hard negatives involving word order shuffling allows it to perform the best. All models suffer from poor object attribute association.}
\label{tab:shuffled}
\end{table}
\begin{table}[t]
\centering
\resizebox{\columnwidth}{!}{  

            \begin{tabular}{lccc}
            \toprule
              Prompt Type & \begin{tabular}{@{}c@{}}EM on \\ \dataname\end{tabular} & \begin{tabular}{@{}c@{}}\funfont{CIC}\\Image \\ score\end{tabular} & \begin{tabular}{@{}c@{}}\funfont{CIC}\\Text \\ score\end{tabular} \\
              \midrule
              Spatial 1-obj L/R & 27.2 & 1.0 & 10.0 \\
              Spatial 2-obj L/R & 0.6 & 4.0 & 10.0 \\
              Temporal & 7.3 & 4.0 & 35.0 \\
              Verb 1-obj & 15.4 & 71.0 & 77.0\\
              Verb 2-obj & 6.6 & 43.0 & 59.0\\
              Adjectives & 38.2 & 74.0 & 92.0\\
              \bottomrule
            \end{tabular}
        }
\caption{Performance of CLIP ViT-B/32 text encoder (\%) on the equivalent prompts in \dataname, and performance of CLIP ViT-B/32 full model on \evalname (\funfont{CIC}). 
% \jack{can we fit \evalname into the headers for the second two columns?} 
% \jack{what are image/text score? same setup as winoground?}
}
\label{tab:downstream_vitb}
\end{table}

\begin{table}[t]
\centering
\resizebox{\columnwidth}{!}{  

            \begin{tabular}{lccc}
            \toprule
              Prompt Type & \begin{tabular}{@{}c@{}}EM on \\ \dataname\end{tabular} & \begin{tabular}{@{}c@{}}\funfont{CIC}\\Image \\ score\end{tabular} & \begin{tabular}{@{}c@{}}\funfont{CIC}\\Text \\ score\end{tabular} \\
              \midrule
              Spatial 1-obj L/R & 26.0 &1.0 & 13.0 \\
              Spatial 2-obj L/R & 15.0 & 3.0 & 13.0 \\
              Temporal & 23.0 & 8.0 & 35.0 \\
              Verb 1-obj & 20.6 & 84.0 & 94.0\\
              Verb 2-obj & 39.4 & 70.0 & 87.0\\
              Adjectives & 42.2 & 95.0 & 92.0\\
              \bottomrule
            \end{tabular}
        }
\caption{Performance of negCLIP ViT-B/32 text encoder (\%) on the equivalent prompts in \dataname, and performance of negCLIP ViT-B/32 full model on \evalname (\funfont{CIC}). 
% \jack{can we fit \evalname into the headers for the second two columns?} 
% \jack{what are image/text score? same setup as winoground?}
}
\label{tab:downstream_negclip}
\end{table}

\begin{table}[t]
\centering
\resizebox{\columnwidth}{!}{  

            \begin{tabular}{lccc}
            \toprule
              Prompt Type & \begin{tabular}{@{}c@{}}EM on \\ \dataname\end{tabular} & \begin{tabular}{@{}c@{}}\funfont{CIC}\\Image \\ score\end{tabular} & \begin{tabular}{@{}c@{}}\funfont{CIC}\\Text \\ score\end{tabular} \\
              \midrule
              Spatial 1-obj L/R & 92.3 & 1.0 & 10.0 \\
              Spatial 2-obj L/R & 28.3 & 2.0 & 20.0 \\
              Temporal & 18.5 & 2.0 & 23.0 \\
              Verb 1-obj & 30 & 75.0 & 91.0\\
              Verb 2-obj & 48.8 & 50.0 & 65.0\\
              Adjectives & 17 & 80.0 & 87.0\\
              \bottomrule
            \end{tabular}
        }
\caption{Performance of RoBERTa-CLIP ViT-B/32 text encoder (\%) on the equivalent prompts in \dataname, and performance of RoBERTa-CLIP ViT-B/32 full model on \evalname (\funfont{CIC}). 
% \jack{can we fit \evalname into the headers for the second two columns?} 
% \jack{what are image/text score? same setup as winoground?}
}
\label{tab:downstream_robertaclip}
\end{table}

\begin{table}[h]
\centering
\resizebox{.9\columnwidth}{!}{  

            \begin{tabular}{lcc}
            \toprule
              Dataset & negCLIP & ChatGPT 2-shot \\
              \midrule
              VG-Relation & 0.81 & 0.90 \\
              VG-Attribution & 0.71 & 0.80 \\
              Flickr30k-PRC & 0.91 & 0.86 \\
              COCO-PRC ViT-B/32 & 0.86 & 0.86 \\
              \bottomrule
            \end{tabular}
        }
\caption{Performance of ChatGPT 2-shot on the ARO benchmark \cite{yuksekgonul2022and}. While the dataset was designed to test VL models' sensitivity to word order shuffling (which is orthogonal to text-only performance on the same data), the textual priors that exist in ARO (e.g., ``horse eating grass'' is more likely than ``grass eating horse'') are less relevant to the probing experiments for \dataname (because the probe must reconstruct \emph{any} given caption in \dataname, including unusual ones, e.g., ``five teenagers riding three butterflies") and do not exist in \evalname due to the paired-image construction.}
\label{tab:aro_text_only}
\end{table}

\end{document}